\documentclass[a4paper,twoside]{article}

\usepackage{epsfig}
\usepackage{subcaption}
\usepackage{calc}
\usepackage{amssymb}
\usepackage{amstext}
\usepackage{amsmath}
\usepackage{amsthm}
\usepackage{multicol}
\usepackage{pslatex}
\usepackage{apalike}
\usepackage{algorithm2e}
\usepackage[bottom]{footmisc}
\usepackage{booktabs}
\usepackage{SCITEPRESS}     

\begin{document}

\title{Barycentric Anchored Weight Search for Robust Model Merging}

\author{\authorname{A. Buades\orcidAuthor{0000-0001-9832-3358} and J.~L.~Lisani\orcidAuthor{0000-0002-7004-2252}}
\affiliation{Universitat de les Illes Balears, IAC3, Spain}
\email{toni.buades@uib.es, joseluis.lisani@uib.es}
}


\abstract{We propose Barycentric Anchored Search, a post-hoc weight-space procedure for improving fine-tuned vision models under distribution shift without additional gradient training or increased inference cost. Given a pre-trained initialization and two checkpoints fine-tuned on a source dataset, the method constructs barycentric candidate weights and performs a small derivative-free search along directions connecting each candidate to the pre-trained model. Candidate weights are selected using a small validation subset of one or more out-of-distribution (OOD) datasets; the selected model is then evaluated on the held-out portions of those datasets, on additional OOD benchmarks, and on the ImageNet validation set to quantify the robustness–accuracy trade-off.
The method is inspired by simplex-based derivative-free optimization, including Nelder–Mead and Torczon multidirectional search, but is specialized to the geometry of neural network weight merging. Unlike Model Soup, which typically benefits from many fine-tuned checkpoints, and unlike Model Stock, which applies a fixed geometric merge, our approach uses only two fine-tuned checkpoints and iteratively adapts the merge using validation feedback from OOD data. Experiments with CLIP ViT-B/32 show that the proposed search substantially improves performance on held-out OOD data, including ImageNet-A, ImageNet-R, and ImageNet-Sketch, while keeping ImageNet accuracy competitive. The resulting weights also improve transfer to OOD datasets not used during candidate selection, outperforming the original fine-tuned checkpoints, naive model soups, and Model Stock in our evaluations. These results suggest that lightweight optimization in anchored weight space provides a practical alternative to fixed weight-averaging rules, geometric merging, or single-model selection.}

\onecolumn \maketitle \normalsize \setcounter{footnote}{0} \vfill

\section{\uppercase{Introduction}}
 
Fine-tuning large-scale pre-trained vision models, such as CLIP \cite{radford2021clip} and {ALIGN} \cite{Jia2021}, has emerged as the standard protocol for transferring general representations to specific downstream tasks. The conventional pipeline involves performing exhaustive hyperparameter sweeps—varying configurations such as learning rates, weight decay, data augmentation strategies, and random seeds—to train multiple independent model checkpoints \cite{wortsman2022soups,jang2025modelstock}. Following this search, practitioners typically select the single best model based on its performance on a held-out validation set and discard the remaining candidates. 

\medskip

Although this strategy is simple, it discards potentially useful information from the remaining checkpoints, which research shows often reside in a common low-error basin of the loss landscape \cite{wortsman2022soups}. Furthermore, standard fine-tuning is prone to a ``robustness gap,'' where improving performance on a target task (In-Distribution) leads to a sharp decline in accuracy on unseen data (Out-of-Distribution) \cite{wortsman2021wiseft}. This trade-off has motivated the development of weight-space combination techniques—such as Model Soups \cite{wortsman2022soups}, {WiSE-FT} \cite{wortsman2021wiseft}, and {Model Stock} \cite{jang2025modelstock}—which aim to leverage the strengths of multiple checkpoints to simultaneously improve accuracy and robustness without increasing computational costs at inference time \cite{wortsman2022soups,jang2025modelstock}.

\medskip

Model Soups averages multiple fine-tuned models and shows that carefully selected soups can outperform the best individual checkpoint without extra inference cost~\cite{wortsman2022soups}. Model Stock pushes this idea further by using only a few fine-tuned models and a layer-wise interpolation rule derived from geometric observations in weight space~\cite{jang2025modelstock}.

\medskip

In this paper, we explore a different route. Instead of directly averaging many models or computing a closed-form interpolation coefficient, we perform a small derivative-free search in weight space. We begin with the pre-trained model $\omega_P$ and a reduced number of fine-tuned models in a source dataset.  We form barycenters of the current candidates, generate nearby anchored perturbations along directions defined by the pre-trained model, and keep only the candidates that improve  accuracy using a small validation subset of one or more out-of-distribution (OOD) datasets. This search is lightweight, requires only model evaluation, and preserves the key practical property of soups and stock: the final predictor is still a single model.

\medskip

Our approach is inspired by simplex-based direct search. In classical Nelder--Mead~\cite{nelder1965simplex}, one refines a simplex by reflection, expansion, contraction, and shrink operations around the centroid of the current vertices. In multidirectional search~\cite{torczon1991convergence,kolda2003direct}, simplex vertices are updated through coordinated expansions and contractions that enjoy a cleaner direct-search interpretation and stronger convergence guarantees than the classical Nelder--Mead heuristic. Our method adapts this intuition to neural network merging: instead of optimizing a low-dimensional continuous vector by repeated simplex deformation, we treat fine-tuned checkpoints and their barycenters as simplex-like vertices in weight space and evaluate anchored moves around them using the pre-trained model as an anchor. The resulting procedure is simple, post-training, and validation-driven.

\medskip

\paragraph{Contributions.}
We make three contributions. First, we propose BAS, a simple barycentric weight-search procedure that starts from a pre-trained checkpoint and two fine-tuned checkpoints, and explores a small set of averaged and anchored candidates without additional training. Second, we interpret BAS through the lens of derivative-free simplex optimization, drawing inspiration from Nelder--Mead and multidirectional search while adapting their geometric ideas to the constraints of neural network weight merging. Third, we provide an empirical comparison with matched baselines, including naive weight averaging and a reproduced Model Stock baseline, and show that BAS can improve OOD accuracy when guided by small validation subsets from the target distribution while largely preserving ImageNet performance.

\section{\uppercase{Related work}}

The primary objective when combining different checkpoints (specifically through weight-space ensembling or averaging) is to simultaneously improve model accuracy and robustness to distribution shifts. Deep ensembling methods \cite{Fort2020Deep,Rane2024Ensemble} are particularly effective since they tend to explore diverse modes in function space rather than clustering within a single mode, but increase computational costs due to the evaluation of several methods. Standard fine-tuning improves accuracy on a target task (In-Distribution) but causes a sharp decline in performance on unseen data (Out-of-Distribution, or OOD)

\medskip

\paragraph{Weight averaging and flat minima.}
Our work is related to stochastic weight averaging, direct weight averaging, and the broader literature on flat minima~\cite{izmailov2018swa,foret2021sam,cha2021swad}. These papers are motivated by the geometric properties of the loss landscape, aiming to reach more optimal regions that individual training runs might miss.  The common intuition is that averaging or interpolating appropriate solutions may move the model toward flatter regions which lead to better generalization because they are more robust to the shift between training and test error surfaces. Our method differs in that it performs a short validation-guided search around barycenters rather than relying on a fixed averaging rule.

\medskip

Within the weight-averaging literature, Model Soup~\cite{wortsman2022soups} is the most directly related baseline to our approach. It showed that averaging multiple fine-tuned models can improve accuracy and robustness without extra inference cost, especially when the checkpoints lie in a shared basin. Their greedy soup procedure sequentially adds candidate models only when validation accuracy improves, thereby avoiding harmful ingredients. This line of work is connected to earlier observations on mode connectivity and the geometry of fine-tuned solutions~\cite{garipov2018mode,frankle2020lmc,neyshabur2020transfer,Fort2020Deep,Nguyen2021When}. 

\paragraph{Robust fine-tuning and anchored interpolation.}
A related line of work interpolates between pre-trained and fine-tuned solutions to improve robustness under distribution shift~\cite{wortsman2021wiseft,kumar2022lpft}. More broadly, model-merging methods based on Fisher weighting or task vectors also combine post-training solutions in parameter space, although typically in settings different from ours~\cite{matena2021fisher,ilharco2023task}. Fast Trainable Projection (FTP) further improves this by efficiently learning per-layer projection constraints to maintain the robustness of the foundation model during transfer \cite{Tian2023Fast}.
CAR-FT \cite{mao2024context} uses zero-shot weights to extract ``context distributions,'' regularizing the fine-tuning process to ensure context-awareness is inherited by the downstream model, thereby improving OOD performance. Supervised contrastive finetuning, such as the FLYP method \cite{Goyal2022Finetune}, mimics the pretraining loss to achieve state-of-the-art robustness across distribution shifts.

\medskip

From the perspective of anchored interpolation, our work is closest in spirit to Model Stock~\cite{jang2025modelstock}, which argued that fine-tuned checkpoints exhibit regular geometric structure in weight space and proposed a layer-wise interpolation between the pre-trained model and a small number of fine-tuned models. Compared with soups, stock aims to reach competitive performance with dramatically fewer fine-tuned checkpoints. In parallel, WiSE-FT showed that interpolating between pre-trained and fine-tuned weights can improve robustness under distribution shift~\cite{wortsman2021wiseft}, reinforcing the role of the pre-trained model as a useful anchor.

\medskip

See \cite{Li2023Deep,Yang2026Model} for a more comprehensive detail on model fusion and merging, and  Gawlikowski et al., 2023 \cite{Gawlikowski2023Survey} for uncertainty.

\section{\uppercase{Method}}

Let $f(x;\omega)$ denote a neural network with parameters $\omega$. We assume access to a pre-trained model $\omega_P$ and {several task-adapted fine-tuned models} $\omega_1, \ldots, \omega_n$ obtained from the same initialization and adapted to the same task. Let $\mathcal{V}$ be a validation set and let
\begin{equation}
F(\omega) = \mathrm{Acc}(f(\cdot;\omega), \mathcal{V})
\end{equation}
be the validation accuracy. Our goal is to obtain a single final checkpoint $\omega^\star$ with improved validation performance In-Distribution and Ouf-Of-Distribution, using only weight-space operations and model evaluation.

\subsection{\uppercase{Barycentric anchored search: general algorithm}}

The method starts with the $n$ vertices $\{\omega_1, \ldots, \omega_n\}$ and the pre-trained vertex $\omega_P$. First we form the barycenter
\begin{equation}
\omega_{n+1} = \frac{\omega_1 + \ldots + \omega_n}{n}.
\end{equation}
For every candidate $\omega_i$ under consideration, $i=1,\ldots,n+1$, we define anchored perturbations
\begin{equation}
\omega_i^{\lambda} = \omega_i + \lambda (\omega_P - \omega_i), \label{eq:perturbation}
\end{equation}
where $\lambda$ controls contraction toward the pre-trained anchor when $\lambda > 0$ and expansion away from it when $\lambda < 0$.
Such perturbations are computed for a set of pre-determined $\lambda \in \Lambda$.

\medskip

Each candidate is updated according to
\begin{equation}
\omega_i \leftarrow \omega_i^{\lambda} \quad \text{if} \quad F(\omega_i^{\lambda}) > F(\omega_i).  \label{eq:update}
\end{equation}

\medskip

The same process is  repeated for $N$ iterations with sets $\Lambda_1, \ldots, \Lambda_N$.  The final model is
\begin{equation}
\omega^\star = \arg\max_{\omega_i} F(\omega_i).
\end{equation}

\subsection{\uppercase{Barycentric anchored search: two refinements. Practical algorithm}}

The method starts with the three vertices $\{\omega_P, \omega_1, \omega_2\}$, although only the two fine-tuned models compete as task-adapted candidates. First we form the barycenter
\begin{equation}
\omega_3 = \frac{\omega_1 + \omega_2}{2}.
\end{equation}
For every candidate $\omega_i$ under consideration, $i=1,2,3$, we define anchored perturbations
\begin{equation}
\omega_i^{\lambda} = \omega_i + \lambda (\omega_P - \omega_i)
\end{equation}
where $\lambda$ controls contraction toward the pre-trained anchor when $\lambda > 0$ and expansion away from it when $\lambda < 0$.

In the first stage we evaluate $\lambda \in \{-0.8, -0.6, -0.4, -0.2, 0.2, 0.4, 0.6, 0.8\}$ and update each candidate according to
\begin{equation}
\omega_i \leftarrow \omega_i^{\lambda} \quad \text{if} \quad F(\omega_i^{\lambda}) > F(\omega_i).
\end{equation}
Then we form a second barycenter
\begin{equation}
\omega_4 = \frac{\omega_1 + \omega_2 + \omega_3}{3},
\end{equation}
and repeat the same anchored perturbations \eqref{eq:perturbation} and updating each candidate \eqref{eq:update}  for $i=1,2,3,4$,  with a smaller step size, now using $\lambda \in \{-0.04, -0.03, -0.02, -0.01, 0.01, 0.02, 0.03, 0.04\}$. The final model is
\begin{equation}
\omega^\star = \arg\max_{\omega_i} F(\omega_i).
\end{equation}

\subsection{\uppercase{Interpretation. Relation to direct-search methods}}

The procedure can be interpreted as a simplex-like search over a tiny set of vertices in weight space. The barycenters play the role of simplex summaries, while anchored perturbations emulate coarse expansion and contraction steps. Unlike Model Soup, which averages many checkpoints, our method keeps the search local and validation-guided. Unlike Model Stock, which computes interpolation ratios analytically, our method uses explicit evaluation to decide whether movement toward or away from the pre-trained anchor is beneficial.

\medskip

This combination operates with very few ingredients. It naturally supports multiple stages with progressively smaller step sizes, which mirrors the coarse-to-fine logic of derivative-free search. 

\medskip

Nelder--Mead  and Torczon's  multidirectional search provide the main algorithmic intuition behind our design. Both methods work with simplex geometry, use only objective evaluations, and improve a current set of candidates through structured geometric transformations~\cite{nelder1965simplex,torczon1991convergence,kolda2003direct}. Our procedure adopts these principles in a form specialized to model merging.

\medskip

The proposed approach is conceptually analogous to the Nelder-Mead algorithm, in which the current fine-tuned checkpoints serve as vertices of a simplex, and the anchored perturbations function as expansion or contraction operations, governed by the sign of $\lambda$. However, whereas the standard Nelder-Mead algorithm iteratively updates only the worst-performing vertex relative to the centroid of the remaining points, our method modifies all checkpoints simultaneously, relative to the pre-trained one.

\medskip

This approach aligns more closely with Torczon’s Multidirectional Search (MDS). Like MDS, our method updates multiple points in parallel rather than relying on a single analytical update. Specifically, both evaluate a structured family of candidate moves around the current geometry and accept only those that improve the validation objective. However, Torczon's algorithm utilizes the best-performing checkpoint as the primary anchorage while we use the pretrained one.

\medskip

Our algorithm differs from both classical methods in an essential way: it does not maintain a full $d+1$-vertex simplex in parameter space. Instead, it searches a restricted weight-space family generated by the pre-trained anchor, the fine-tuned checkpoints, and their barycenters. 
For such a reason, we do not claim any convergence properties of the proposed method to any minimum.  However, following classical analysis on step size control for direct search methods on optimization \cite{audet2006mesh,dolan2003local}, we reduce the magnitude of the tested step sizes at each iteration  $\Lambda_1, \ldots, \Lambda_N$.

\begin{figure}
\begin{center}
\fbox{%
\parbox{\linewidth}{%
\small
\textbf{Algorithm 1: Barycentric anchored weight search (BAS)}
\begin{enumerate}
    \item Start from pre-trained weights and two fine-tuned checkpoints: $\{\omega_P, \omega_1, \omega_2\}$.
    \item Compute the barycenter $\omega_3 = (\omega_1 + \omega_2)/2$.
    \item For each candidate $\omega_i$, evaluate 
    
    $\omega_i^{\lambda} = \omega_i + \lambda(\omega_P - \omega_i)$ 
    
    with 
    
    $\lambda \in \{-0.8, -0.6, -0.4, -0.2, 0.2, 0.4, 0.6, 0.8\}$.
    \item If $F(\omega_i^{\lambda}) > F(\omega_i)$, replace $\omega_i$ by the improved candidate.
    \item Compute $\omega_4 = (\omega_1 + \omega_2 + \omega_3)/3$.
    \item Re-evaluate anchored perturbations using 
    
    $\lambda \in \{-0.04, -0.03, -0.02, -0.01, 0.01, 0.02, 0.03, 0.04\}$.
    \item Keep the final checkpoint $\omega^\star = \arg\max_{\omega_i} F(\omega_i)$.
\end{enumerate}}
}
\end{center}
\caption{BAS procedure using one pre-trained checkpoint and two fine-tuned checkpoints.}
\label{alg:bas}
\end{figure}

\section{\uppercase{Experiments}}

We evaluate Barycentric Anchored Search (BAS) in the same post-hoc model-merging regime considered by Model Soups and Model Stock: all compared methods produce a single checkpoint and therefore have the same inference cost as an individual model.  The goal of the experiments is not to introduce a new fine-tuning recipe, but to test whether a small validation-guided search in the weight space spanned by a pre-trained model and two fine-tuned checkpoints can improve robustness under natural distribution shift.

\subsection{\uppercase{Experimental protocol}}

We use the CLIP ViT-B/32 architecture.  The pre-trained checkpoint and two ImageNet fine-tuned checkpoints were downloaded from the public repository accompanying Model Stock.  Thus, all methods compared in this section start from the same three weights: the pre-trained model $\omega_P$ and two independently fine-tuned ImageNet checkpoints, $\omega_1$ and $\omega_2$.  BAS performs no gradient-based optimization and does not fine-tune any additional model.  It only evaluates the candidates generated by Algorithm~\ref{alg:bas} and selects the candidate with the highest validation accuracy.

We consider one in-distribution benchmark, ImageNet, and three natural out-of-distribution benchmarks: ImageNet-A, ImageNet-R, and ImageNet-Sketch.  For each dataset, we construct two disjoint subsets: a validation subset used only to guide BAS, and a test subset used only for the final accuracies reported in Table~\ref{tab:main_results}.  For ImageNet, we follow the split used by Model Soups for greedy soup construction: approximately $2\%$ of the ImageNet training set is used as validation data, while the official ImageNet validation set is used as test data.  For ImageNet-A, ImageNet-R, and ImageNet-Sketch, we randomly sample $20\%$ of the images for validation and use the remaining $80\%$ for testing.

The second column of Table~\ref{tab:main_results} specifies which validation subset is used by BAS.  For example, ``BAS / ImageNet-A'' means that the BAS candidate updates and final checkpoint selection are performed using only the ImageNet-A validation split.  The resulting single checkpoint is then evaluated on all test sets: ImageNet, ImageNet-A, ImageNet-R, and ImageNet-Sketch.  This protocol lets us measure both targeted OOD adaptation and cross-OOD transfer.

\paragraph{Implementation and compute.}
All computations were performed in Google Colab notebooks.  Our notebooks were based on the public example notebooks released with the Model Soups and Model Stock repositories, and were adapted to evaluate BAS candidates, the naive soup, and the Model Stock baseline under the same splits.  Since BAS is a post-hoc procedure, the reported experiments do not train new neural networks; the computational cost comes from loading the public CLIP ViT-B/32 checkpoints, forming candidate weight averages, and evaluating them on the validation and test subsets.  
Upon acceptance, we will release a public GitHub repository containing the code and Colab notebooks needed to reproduce these experiments.

\paragraph{Baselines.}
We compare BAS with five reference checkpoints.  The first three are the pre-trained CLIP ViT-B/32 model and the two fine-tuned checkpoints used as BAS inputs.  The fourth baseline is a naive two-model soup, defined as the arithmetic average $(\omega_1+\omega_2)/2$.  The fifth baseline is Model Stock computed from the same pre-trained model and the same two fine-tuned checkpoints.  These baselines separate three effects: supervised fine-tuning, unvalidated barycentric averaging, and the geometry-based anchored merge of Model Stock.  BAS uses the same available weights but additionally performs a small validation-driven anchored search.


\begin{table*}[t]
\centering
\small
\caption{Top-1 accuracy of the proposed BAS method and matched baselines.  The second column indicates the validation set used to guide BAS.  All accuracies are computed on held-out test sets.  The last column reports the average accuracy over the three OOD test sets: ImageNet-A, ImageNet-R, and ImageNet-Sketch.  The best result in each column is shown in bold and the second-best result is underlined.}
\label{tab:main_results}
\resizebox{\linewidth}{!}{%
\begin{tabular}{lcccccc}
\toprule
Method & Validation set & ImageNet & ImageNet-A & ImageNet-R & ImageNet-Sketch & OOD Avg. \\
\midrule
Pre-trained & --  & 0.6334 & 0.3178 & 0.6772 & 0.4223 & 0.4724 \\
Fine-tuned 1 & --  & 0.7820 & 0.2384 & 0.5869 & 0.4135 & 0.4129 \\
Fine-tuned 2 & --  & 0.7840 & 0.2503 & 0.5918 & 0.4138 & 0.4186 \\
\midrule
Naive Soup & -- & 0.7897 & 0.2595 & 0.6133 & 0.4345 & 0.4358 \\
Model Stock & -- & \textbf{0.7967} & 0.3239 & 0.6710 & 0.4706 & 0.4885 \\
\midrule
BAS & ImageNet & \underline{0.7946} & 0.3058 & 0.6549 & 0.4640 & 0.4749 \\
BAS & ImageNet-A & 0.7635 & \underline{0.3571} & \textbf{0.7150} & \underline{0.4828} & \underline{0.5183} \\
BAS & ImageNet-R & 0.7131 & \textbf{0.3580} & \textbf{0.7150} & 0.4678 & 0.5136 \\
BAS & ImageNet-Sketch & 0.7721 & \underline{0.3571} & \underline{0.7133} & \textbf{0.4870} & \textbf{0.5191} \\
\bottomrule
\end{tabular}
}
\end{table*}

\subsection{\uppercase{Relation to Model Soups and Model Stock}}

Model Soups showed that averaging independently fine-tuned models can improve both in-distribution and distribution-shift accuracy without increasing inference cost, especially when the checkpoints lie in a common low-error basin~\cite{wortsman2022soups}.  Their strongest greedy soups, however, are typically constructed from a large pool of fine-tuned checkpoints and use a validation set to decide which ingredients to include.  Model Stock was proposed as a more economical alternative: it uses only a few fine-tuned models and exploits the pre-trained checkpoint as a geometric anchor~\cite{jang2025modelstock}.

BAS occupies an intermediate position.  Like Model Stock, it uses only two fine-tuned checkpoints and the pre-trained model.  Like greedy Model Soups, it uses validation accuracy to decide which weight-space candidate should be kept.  The main difference is that BAS does not merely accept or reject full checkpoints as soup ingredients.  Instead, it constructs barycenters and evaluates local anchored contractions and expansions around each candidate.  This makes the method especially suitable when only a very small number of fine-tuned models is available, but a small validation subset from the target or deployment distribution can be obtained.

The protocol is also intentionally stricter than simply selecting the best candidate on the final test sets.  The BAS validation subsets are disjoint from the test subsets, and the reported values in Table~\ref{tab:main_results} are always computed on held-out data.  Therefore, the improvements should be interpreted as validation-guided OOD adaptation rather than test-set oracle selection.

\subsection{\uppercase{Results and discussion}}

The first observation is that both post-hoc merging baselines improve the individual fine-tuned checkpoints.  The naive soup improves ImageNet accuracy from $0.7840$ to $0.7897$ and also improves the three OOD datasets relative to the two fine-tuned checkpoints.  Model Stock is stronger: it obtains the best ImageNet accuracy, $0.7967$, and also improves over the fine-tuned checkpoints on ImageNet-A, ImageNet-R, and ImageNet-Sketch.  These results are consistent with the central message of Model Soups and Model Stock: averaging or anchored merging can recover part of the robustness lost during supervised fine-tuning while preserving the single-model inference cost.

BAS guided by ImageNet validation gives the most conservative behavior.  It reaches $0.7946$ on ImageNet, only $0.0021$ below Model Stock and above all other baselines.  Its OOD accuracy is also substantially better than the two fine-tuned checkpoints and the naive soup, although below Model Stock on the three OOD benchmarks.

The most interesting behavior appears when BAS is guided by a small OOD validation subset.  Using ImageNet-A validation improves ImageNet-A accuracy to $0.3571$, compared with $0.3239$ for Model Stock, $0.2595$ for the naive soup, and at most $0.2503$ for the two fine-tuned checkpoints.  The same selected model also reaches $0.7150$ on ImageNet-R and $0.4828$ on ImageNet-Sketch, showing that the gain is not restricted to the dataset used for validation.  BAS guided by ImageNet-R gives a very similar result: it obtains the best ImageNet-A accuracy, $0.3580$, and ties the best ImageNet-R accuracy, $0.7150$.  BAS guided by ImageNet-Sketch obtains the best ImageNet-Sketch accuracy, $0.4870$, while remaining competitive on ImageNet-A and ImageNet-R.

Averaging the accuracy over the three OOD test sets highlights the same trend.  The pre-trained model obtains an average OOD accuracy of $0.4724$, while the two fine-tuned checkpoints drop to approximately $0.413$.  The naive soup improves this average to $0.4358$, and Model Stock reaches $0.4885$.  BAS guided by ImageNet-A, ImageNet-R, and ImageNet-Sketch obtains average OOD accuracies of $0.5183$, $0.5136$, and $0.5191$, respectively.  Thus, a small validation subset from any of the considered OOD datasets is sufficient to select a BAS checkpoint that outperforms all baselines on average OOD accuracy.

The cost of this improved robustness is a controlled reduction in ImageNet accuracy.  With ImageNet-A or ImageNet-Sketch validation, BAS obtains ImageNet accuracies of $0.7635$ and $0.7721$, respectively.  These values are below Model Stock but remain close to the original fine-tuned checkpoints, especially considering the large OOD gains.  The ImageNet-R-guided model shows the largest ImageNet drop, $0.7131$, suggesting that ImageNet-R validation can select a more robustness-oriented point in the searched weight-space family. 

Overall, the results support three conclusions.  First, BAS can exploit small OOD validation sets to move beyond fixed averaging and fixed geometric interpolation rules.  Second, the resulting model often transfers across OOD datasets: validation on ImageNet-A, ImageNet-R, or ImageNet-Sketch improves not only the corresponding dataset but also other distribution shifts.  Third, the method preserves the practical advantages of Model Soups and Model Stock: once the search is complete, deployment uses a single set of weights and therefore incurs no ensemble-time memory or compute overhead.

\section{\uppercase{Conclusions}}

We introduced Barycentric Anchored Search, a lightweight post-hoc method for merging fine-tuned models in weight space.  Starting from a pre-trained checkpoint and two fine-tuned checkpoints, BAS constructs barycentric candidates and performs a short validation-guided search along anchored directions connecting each candidate to the pre-trained model.  The method is inspired by derivative-free simplex search, but is adapted to the structure of fine-tuned neural networks and the practical constraints of model merging.

The experimental results show that BAS is particularly effective when a small validation subset from an OOD distribution is available.  In this setting, BAS consistently improves the corresponding OOD test accuracy and, more importantly, often transfers to other OOD datasets.  Across ImageNet-A, ImageNet-R, and ImageNet-Sketch, BAS guided by any one of these OOD validation sets achieves higher average OOD accuracy than the pre-trained model, the two fine-tuned checkpoints, the naive soup, and Model Stock.  At the same time, the final predictor remains a single model, so the gains do not come from ensembling or increased inference-time cost.

The results also clarify the role of the validation set.  When BAS is guided by ImageNet validation, it behaves conservatively and preserves ImageNet accuracy, matching the objective used for selection.  When BAS is guided by OOD validation, it moves to a different point in the anchored barycentric family, sacrificing some ImageNet accuracy in exchange for substantially better robustness.  This behavior suggests that BAS can be used as a practical model-selection and adaptation tool: the same small set of precomputed weights can be adapted to different deployment priorities simply by changing the validation distribution used to guide the search.

Several limitations remain.  The present study uses two fine-tuned checkpoints and a small number of anchored step sizes; larger checkpoint sets, deeper multistage schedules, and adaptive choices of $\lambda$ may further improve the trade-off between in-distribution and OOD performance.  We have also focused on CLIP ViT-B/32 and ImageNet-derived distribution shifts.  Future work should evaluate the method on larger architectures, other pre-training families, additional OOD benchmarks, and non-vision tasks.  Finally, a more systematic analysis of the geometry of BAS-selected checkpoints could help explain why validation on one OOD dataset often improves accuracy on others.

Despite these open questions, the current results indicate that a very small amount of validation feedback, combined with a constrained barycentric search around pre-trained and fine-tuned weights, is a powerful mechanism for robust model merging.  BAS therefore complements both Model Soups and Model Stock: it retains their single-model deployment advantage while adding an explicit validation-driven mechanism for adapting merged weights to target distribution shifts.

\section*{Acknowledgments and Disclosure of Funding}
Acknowledgments and funding information are omitted for anonymous review.

\bibliographystyle{apalike}
{
\small
\bibliography{references}
}

\end{document}